\documentclass[10pt]{article}

\usepackage{bm}
\usepackage{bbm}
\usepackage{fullpage}  

\usepackage{url}
\usepackage{algpseudocode}
\usepackage{multirow} 
\usepackage{hhline}  
\usepackage{verbatim}
\usepackage{graphicx}
\usepackage{amsmath,amssymb,amsfonts,graphicx,amsthm,mathtools,nicefrac}
\usepackage{lscape}
\usepackage{color}
\usepackage{authblk}
\usepackage{subfigure}
\usepackage[ruled,vlined]{algorithm2e}

\DeclareMathOperator*{\argmin}{arg\,min}
\usepackage{bbding}
\usepackage{indentfirst}
\usepackage{graphicx}

\usepackage{multirow}
\allowdisplaybreaks
\newcommand\numberthis{\addtocounter{equation}{1}\tag{\theequation}}

\newtheorem{theorem}{Theorem}[section]
\newtheorem{assumption}{Assumption}[section]
\newtheorem{remark}{Remark}[section]

\numberwithin{equation}{section}

\newtheorem{lemma}[theorem]{Lemma}

\usepackage{bbm}

\usepackage{amsmath,mathrsfs,amsfonts,amssymb} 
\usepackage[dvipsnames]{xcolor}
\usepackage{subfigure}
\usepackage{parskip}

\usepackage[utf8]{inputenc}
\usepackage[T1]{fontenc}
\usepackage{hyperref}
\usepackage{url}
\usepackage{booktabs}
\usepackage{amsfonts}
\usepackage{nicefrac}
\usepackage{microtype}
\usepackage{xcolor}
\usepackage{graphicx}
\usepackage{subcaption}
\usepackage{amsmath,amsthm,amssymb}
\usepackage{lipsum}
\usepackage[export]{adjustbox}
\usepackage{enumitem}
\usepackage{wrapfig,lipsum,booktabs}

\newcommand{\mycommentstyle}[1]{\color[HTML]{0671b9}{\small #1}}
\SetKwComment{Comment}{\mycommentstyle{// }}{}
\usepackage{color, colortbl}
\definecolor{LightCyan}{rgb}{0.78,0.92,1}
\definecolor{IMPROVE}{rgb}{0.043,0.396,0.101}

\begin{document}

\title{Improving Multi-Step Reasoning Abilities of Large Language Models with Direct Advantage Policy Optimization}

\author[1,2]{Jiacai Liu}
\author[1*]{Chaojie Wang}
\author[1]{Chris Yuhao Liu}
\author[1]{Liang Zeng}
\author[1]{Rui Yan}
\author[1,2]{Yiwen Sun}  
\author[1]{\\Yang Liu} 
\author[1]{Yahui Zhou}

\affil[1]{Skywork AI, Kunlun Inc.}
\affil[2]{Fudan University.}

\date{}
\maketitle

\begin{abstract}
The role of reinforcement learning (RL) in enhancing the reasoning of large language models (LLMs) is becoming increasingly significant. Despite the success of RL in many scenarios, there are still many challenges in improving the reasoning of LLMs. One challenge is the sparse reward, which makes optimization difficult for RL and necessitates a large amount of data samples. Another challenge stems from the inherent instability of RL, particularly when using Actor-Critic (AC) methods to derive optimal policies, which often leads to unstable training processes.  To address these issues, we introduce Direct Advantage Policy Optimization (DAPO), an novel step-level offline RL algorithm. Unlike standard alignment that rely solely  outcome rewards to optimize policies (such as DPO),  DAPO employs a critic function to predict the reasoning accuracy at each step, thereby generating dense signals to refine the generation strategy. Additionally, the Actor and Critic components in DAPO are trained independently, avoiding the co-training instability observed in standard AC algorithms like PPO.  We train DAPO on mathematical and code query datasets and then evaluate its performance on multiple benchmarks. Our results show that DAPO can effectively enhance the mathematical and code capabilities on both SFT models and RL models, demonstrating the effectiveness of DAPO.
\let\thefootnote\relax\footnotetext{Work done as an intern at Skywork AI, Kunlun Inc.}
\let\thefootnote\relax\footnotetext{* Corresponding author.}

\noindent
\end{abstract}

\section{Introduction}
In the rapidly evolving landscape of artificial intelligence, large language models (LLMs) have emerged as a cornerstone of natural language processing (NLP) and beyond. These models, trained on vast corpora of text data, have demonstrated an unprecedented ability to understand \cite{diag_gen,question_answer}, generate and reasoning such as solving mathematical problems \cite{MetaMath,DeepSeekMath,skywork_math,Q_star_chaojie} and code generations \cite{swe-bench,Chen2023TeachingLL}. When the token generation process of a LLM is modeled as a Markov Decision Process (MDP), it can be naturally optimized and aligned with human preference using reinforcement learning (RL) methods, known as Reinforcement Learning from Human Feedback (RLHF). The pipline of RLHF generally consists of two stages: 1) Train a reward model (often under Bradley-Terry \cite{Bradley_Terry} model) to label the responses generated by the LLM. 2) Sample a lot of responses and use RL methods to optimize LLM's generation policy by these responses and the associated rewards.

Despite the success of RLHF in various fields \cite{RAFT,DPO,DRO,MathCoder,PPO,RLHF_safety},  it still encounters challenges and difficulties in the field of reasoning. One of the key challenges is the sparsity of rewards \cite{sparse_reward_1,sparse_reward_2}. When use LLMs for mathematical problem-solving and code generation, rewards are only assigned to the terminal tokens. This implies that the generation of intermediate tokens receives no direct reward, and the optimization direction relies solely on the backpropagation of the reward from the terminal token. Consequently, given the vast generation space, the successful training of LLMs requires extensive sampling to address exploration challenges, posing a significant challenge for on-policy methods like PPO \cite{PPO}, which have high sampling costs and low sample efficiency, especially under limited computational resources. Another key challenge is the unstable training process when applying standard actor-critic methods \cite{PPO,SAC,TD3} to LLMs especially under limited computational resources. These methods train the actor and critic simultaneously and the regression targets of both actor and critic are non-stationary and change in response to the updates of the other during the training process. This interwoven update mechanism often leads to an unstable training process that is prone to collapse \cite{Deadly_Triad,TD3} especially when the critic is not a good approximation of the true value function.

In order to overcomes these challenges,  we introduce Direct Advantage Policy Optimization (DAPO), a new step-level offline RL method designed to enhance LLM's performance in reasoning tasks. In order to overcome the challenge of sparse rewards, DAPO use a critic function to estimate the accuracy of each reasoning step, creating the dense signals for the policy optimization. Compared with standard actor-critic methods that train the actor and critic simultaneously, DAPO trains the critic  firstly to ensure it's a good approximation of true value function then optimize the actor using the advantage dataset produced by the pretrained critic. The procedure of DAPO is visualized in Figure. We summarize our contributions as follows:
\begin{itemize}
    \item We introduce DAPO, a new step-level, offline RL method that learns from all generated samples. DAPO optimize the generation of the reasoning steps by sampling multiple candidate steps for each intermediate step $s$ and updating the policy at $s$ according to the advantages of each candidate steps.
    
    \item We give theoretical analysis for the optimization problem of DAPO. Our main theoretical result (see Theorem \ref{theorem:monotonic improvement}) ensures that DAPO will produce a better policy on training distribution     if certain conditions are meet. Besides, our theoretical result also implies that DAPO can always learn a better policy until the reference policy is already optimal.
    
    \item We conduct extensive experiments to demonstrate the effectiveness of DAPO. DAPO improves the performance on multiple  mathematical and coding benchmarks even on the state-of-art 7B RL model, qwen-2.5-math-instruct. Our experiments also show that iteraitve DAPO can further improves the performance. Compared with standard actor-critic methods, DAPO stabilizes the training of process significantly by splitting the training actor and critic into two individual stages.

\end{itemize}

\begin{figure}[t]
    \centering
    \includegraphics[width=1.05\linewidth]{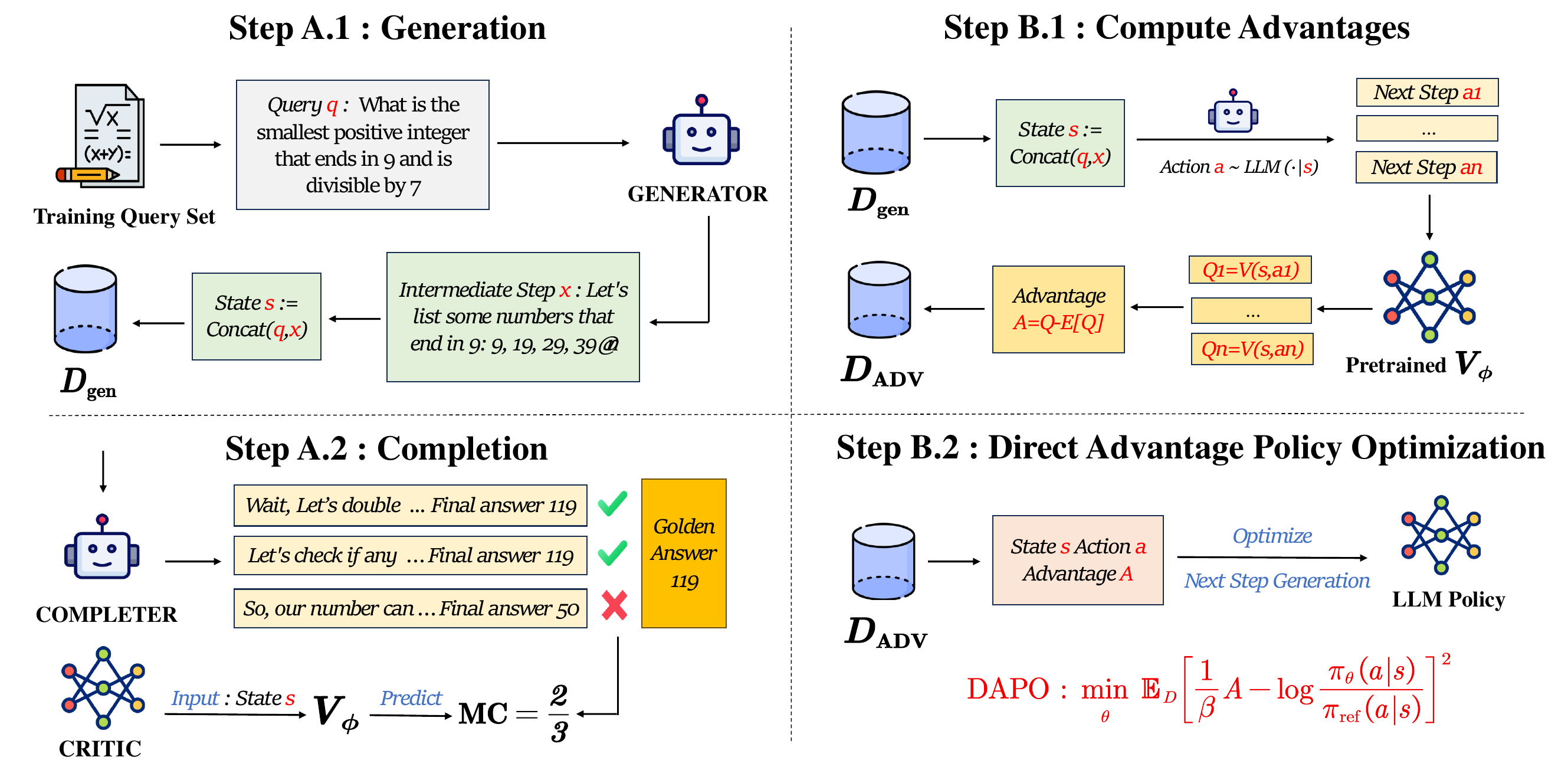}
    \caption{\textbf{Direct Advantage Policy Optimization (DAPO).} The whole training procedure of DAPO consists of two individual stages : 1) \textbf{Critic Training (left)}. Given the training query set $\mathcal{Q}$, for each query $q \in \mathcal{Q}$, DAPO uses the generator $\pi_{\text{gen}}$ to generate multiple rollouts from $q$ and derive a dataset of training states denoted as $\mathcal{D}_{\text{gen}}$. For each state in $s \in \mathcal{D}_{\text{gen}}$,  DAPO uses the completer (often $\pi_{\text{ref}}$) to sample multiple sub-trajectory from $s$ and 
    collects the MC value as the estimation of the true value. Then a critic network $V_\phi$ is trained to approximate the value function of the completer. 2) \textbf{Policy Optimization (right)}. After extracting multiple next steps $\left\{ a_i \right\} _{i=1}^{n}$ of each state $s$ from the completions in step 2, DAPO then uses the trained critic to compute the advantages for all actions by $\forall i \in [n] : A_i=Q_i-\frac{1}{n}\sum_{j=1}^n{Q_j}$ and $Q_i$ is the predicted value of Concat($s,a_i$). Finally, DAPO fits the policy ratio to the advantage of each state-action pair to optimize the generation of reasoning steps.}
    \label{fig:enter-label}
\end{figure}

\subsection{Related Work}
\paragraph{RLHF} Generally speaking, RLHF solves the problem formulated as 
\begin{align}
\underset{\theta}{\max}\,\, \mathbb{E} _{x\sim \mathcal{D}}\mathbb{E} _{\forall t:a_t\sim \pi _{\theta}\left( \cdot |s_t \right)}\left[ \sum_{t=0}^{T-1}{r\left( s_t,a_t \right) -\beta
\mathrm{KL}\left( \pi _{\theta}\left( \cdot |s_t \right) ,\pi _{\mathrm{ref}}\left( \cdot |s_t \right) \right)}|s_0=s \right],
\label{RLHF}
\end{align}
where $\mathcal{D}$ is the distribution of training prompts $x$, 
$a_t$ is the action (or the tokens generated) at time $t$, $T$ is the decision horizon,  $r(\cdot)$ is the ground truth reward function, $\pi_\theta$ is the parameterized generation policy of LLM, $\pi_\text{ref}$ is the initial reference policy, $\beta$ is the KL coefficient. We present a more formal formulation of RLHF problem in Section \ref{Preliminary}. Notice that the definition on what is a single action can be very flexible, thus different definition of a action yields different algorithm designs and theoretical results. Response-level RLHF regards a full response as a single action, and the RLHF problems \eqref{RLHF} reduces to a bandit problem \cite{RLHF}, i.e.,
\begin{align}
\underset{\theta}{\max}\,\,\mathbb{E} _{x\sim \mathcal{D}}\left[ \mathbb{E} _{y\sim \pi _{\theta}\left( \cdot |s \right)}\left[ r\left( x,y \right) \right] -\mathrm{KL}\left( \pi _{\theta}\left( \cdot |x \right) ,\pi _{\mathrm{ref}}\left( \cdot |x \right) \right) \right],
\label{response-level RLHF}
\end{align}
where $y$ is the response of $x$ outputted by the LLM. A great advantage of considering the RLHF problem in the bandit setting is that the optimal policy $\pi^*_\beta$ of \eqref{response-level RLHF} has the analytic solution given by 
\begin{align}
\forall x,y:\log \frac{\pi _{\beta}^{*}\left( y|x \right)}{\pi _{\mathrm{ref}}\left( y|x \right)}=\frac{1}{\beta}\left( r\left( x,y \right) -\underset{:=\hat{V}_{\mathrm{ref}}\left( x \right)}{\underbrace{\beta \log \left\{ \mathbb{E} _{y^{\prime}\sim \pi _{\mathrm{ref}}\left( \cdot |x \right)}\left[ \exp \left\{ \frac{1}{\beta}r\left( x,y^{\prime} \right) \right\} \right] \right\} }} \right).
\label{eq:bellman_eq in response level}
\end{align}
In order to solve $\pi^*_\beta$, DPO \cite{DPO} use the Bradley-Terry model to parametrize the reward model $r$ and infers the optimal policy directly from pair-wise data  without accessing to  $\hat{V}_{\text{ref}}$. DRO \cite{DRO} takes a straight way and solves the optimal policy directly by  
\begin{align*}
\underset{\theta}{\min}\,\, \mathbb{E}_{x,y} \left[ \left( \log \frac{\pi _{\theta}\left( y|x \right)}{\pi _{\mathrm{ref}}\left( y|x \right)}-\frac{1}{\beta}\left( r\left( x,y \right) -\hat{V}_{\mathrm{ref}}\left( x \right) \right) \right)^2 \right].
\end{align*}

However, when an action $a$ is defined as a token or a reasoning step, the decision horizon $T$ of RLHF problem \eqref{RLHF} is generally larger than one. In this case, by Theorem \ref{theorem:optimal policy}, the optimal policy $\pi^*_\beta$ satisfies
\begin{align}
\log \frac{\pi _{\beta}^{*}\left( a|s \right)}{\pi _{\mathrm{ref}}\left( a|s \right)}=\frac{1}{\beta}\left( Q_{\beta}^{*}\left( s,a \right) -V_{\beta}^{*}\left( s \right) \right)\quad \text{and} \quad Q_{\beta}^{*}\left( s,a \right)=r\left( s,a \right) +V_{\beta}^{*}\left( s^{\prime} \right),
\label{eq:bellman_eq}
\end{align}
where $V^*_\beta$ is the value function of $\pi^*_\beta$ defined in \eqref{def:value_function}, $s$ is the current state, $a$ is the action, $s^\prime = f(s,a)$ is the successive state. See Section \ref{Preliminary} for more information. Several policy gradient ascent \cite{pg} based methods are applied to solve the optimal policy. PPO use GAE 
\cite{GAE} to estimate the policy gradient and a clip trick to keep the training process stable. Considering the computation efficiency, GRPO \cite{DeepSeekMath}, RLOO \cite{RLOO}, Remax \cite{ReMax} remove the critic and use the sampling rewards to estimate the gradient. \cite{step_DPO,step_DPO_2,step_DPO_3} adapt the DPO algorithm to the step-level setting, developing a series of methods known as step-DPO. These methods also uses the BT model to parametrize the reward model and infer the optimal policy from \eqref{eq:bellman_eq} without the knowledge of $V^*_\beta$. However, these methods still requires the collection of massive pairwise trajectory data. DQO takes the standard actor-critic route like PPO and solves the optimal policy using the Soft Actor-Critic \cite{SAC}(SAC)  Method.

\section{Preliminaries}
\label{Preliminary}
In this section, we introduce the mathematical formulation of step-level RLHF problem and some important theoretical results. 

\paragraph{RLHF on step-level MDP}  Compared with standard RLHF, which regard outputting a full response as a single action, step-level RLHF treats outputing a full reasoning step as a single action. More Specifically, the fundamental model of step-level RLHF can be represented by a tuple $\mathcal{M}=(\mathcal{S},\mathcal{A},r,f,\mu)$, where $\mathcal{S}$ is the state space and a state $s \in \mathcal{S}$ is any reasoning steps outputted by the LLM so far, $\mathcal{A}$ is the action space and an action $a \in \mathcal{A}$ is any valid singular reasoning step, $r(s,a)$ is the step-wise reward,  $f$ is the deterministic transition model between steps $(i.e., \forall t \in \mathbb{N}, s_{t+1} = f(s_t, a_t) $), $\mu$ is the distribution of the initial state $s_0$. Under this setting, the objective of RLHF is to find a policy that maximize the cumulative reward in the trust region of reference policy $\pi_\text{ref}$,
\begin{align}
\pi _{\beta}^{*}\in \underset{\pi \in \Pi}{\mathrm{arg}\max}\,\,\mathbb{E} _{\forall t\in \mathbb{N} : a_t\sim \pi \left( \cdot |s_t \right)}\left[ \sum_{t=0}^{T-1}{r\left( s_t,a_t \right)}-\beta \cdot \mathrm{KL}\left( \pi \left( \cdot |s \right) ||\pi _{\mathrm{ref}}\left( \cdot |s \right) \right) |s_0\sim \mu \right],
\label{objective}
\end{align}
given the KL coefficient $\beta$, and the set of all admissible policies $\Pi$. Here $T$ is the random decision horizon (i.e., LLM outputs a full response after $T$ reasoning steps and $s_T$ is a terminate state that either contains the [EOS] token or is truncated due to the length limit). For easy of notation, we define $\mathbb{P}^\pi$ as the probability function of the trajectory $(s_0,a_0,...,a_{T-1})$ induced by the policy $\pi$ and the expectation $
\mathbb{E} ^{\pi}_\rho\left[ \cdot \right]$ as 
$$
\mathbb{E} _{\rho}^{\pi}\left[ g\left( s,a \right) \right] :=\mathbb{E} _{\mathbb{P} ^{\pi}}\left[ \sum_{t=0}^{T-1}{g\left( s_t,a_t \right)}|s_0\sim \rho \right] 
$$
for any real-valued function $g:\mathcal{S}\times\mathcal{A} \rightarrow \mathbb{R}$ and $\rho \in \Delta(\mathcal{S})$.

\begin{assumption}We assume there exists a constant $H_1 < \infty$ such that for any policy $\pi \in \Pi$ and initial state $s_0 \in \mathcal{S}$, the random decision horizon T satisfies $T<H_1 \,\, $ almost surely. We also assume that there exists a constant $H_2 < \infty$ such that for any singular reasoning step $a \in \mathcal{A}$, the token length of $a$ 
is no greater than $H_2$ so that the state space $\mathcal{S}$ and action space $\mathcal{A}$ are both finite. 
\label{assumption:T is finite}
\end{assumption}

\begin{assumption}[Traversal initial distribution] We assume for any state $s \in \mathcal{S}$, $\mu(s) > 0$
\label{assumption: mu > 0}
\end{assumption}
\begin{remark} Assumption \ref{assumption: mu > 0} is made just for the simplicity of theoretical analysis and  is commonly appeared in the previous works on the convergence analysis of policy optimization like \cite{Agarwal_pg_2019,mei_softmax_pg,ppgliu,pg-liu,hadamard_pg,mei2021normalized}.
\end{remark}

\paragraph{Value functions} Given any state $x \in \mathcal{S}$, we define the KL-constrained state value function as
\begin{align}
V_{\beta}^{\pi}\left( s \right) :=\mathbb{E} _{\forall t\in \mathbb{N} :a_t\sim \pi \left( \cdot |s_t \right)}\left[ \sum_{t=0}^{T-1}{\left( r\left( s_t,a_t \right) -\beta \mathrm{KL}\left( \pi \left( \cdot |s_t \right) ,\pi _{\mathrm{ref}}\left( \cdot |s_t \right) \right) \right)}|s_0=s \right],
\label{def:value_function}
\end{align}
and the KL-constrained  state-action value function as 
\begin{align}
Q_{\beta}^{\pi}\left( s,a \right) :=r\left( s,a \right) + V_{\beta}^{\pi}\left( f(s,a) \right)
\label{def:Q_function}.
\end{align}
The advantage function $A^\pi_\beta$ is defined as
\begin{align}
A_{\beta}^{\pi}\left( s,a \right) :=Q_{\beta}^{\pi}\left( s,a \right) -V_{\beta}^{\pi}\left( s \right) -\beta \log \frac{\pi \left( a|s \right)}{\pi _{\mathrm{ref}}\left( a|s \right)}.
\label{def:advantage_function}
\end{align}
We also define the optimal value function as
\begin{align*}
\forall s\in \mathcal{S} :  V_{\beta}^{*}\left( s \right) =\underset{\pi \in \Pi}{\max}\,\,V_{\beta}^{\pi}\left( s \right),
\end{align*}
and $Q_{\beta}^{*}\left( s,a \right) =r\left( s,a \right) +V_{\beta}^{*}\left( s \right)$. 
The objective \eqref{objective} can be rewritten as 
$$
\pi _{\beta}^{*}\in \underset{\pi \in \Pi}{\mathrm{arg}\max}\,\,\left\{ V_{\beta}^{\pi}\left( \mu \right) :=\mathbb{E} _{s_0\sim \mu}\left[ V_{\beta}^{\pi}\left( s_0 \right) \right] \right\}.
$$
For unregularized state value functions, i.e. $\beta=0$, we use $V^{\pi},Q^{\pi},A^{\pi}$ instead of $
V_{0}^{\pi},Q_{0}^{\pi},A_{0}^{\pi}$ to be aligned with the standard RL literatures.

\paragraph{Bellman Operators} For arbitrary policy $\pi\in \Pi$ and any value function $V:\mathcal{S} \rightarrow \mathbb{R}$, the Bellman operator $\mathcal{T}^{\pi}_\beta$ is defined as
\begin{align*}
\forall s \in \mathcal{S} : [\mathcal{T} _{\beta}^{\pi}V]\left( s \right) :=
\mathbb{E} _{a\sim \pi \left( \cdot |s \right)}\left[ r\left( s,a \right) +V\left( f(s,a) \right) -\beta \log \frac{\pi \left( a|s \right)}{\pi _{\mathrm{ref}}\left( a|s \right)} \right].
\numberthis
\label{def:Bellman_operator}
\end{align*}
The Bellman optimality operator $\mathcal{T} _{\beta}$ is defined as
\begin{align*}
\forall s\in \mathcal{S} :[\mathcal{T} _{\beta}V]\left( s \right) &:=\underset{\pi \in \Pi}{\max}\,\,\left[ \mathcal{T} _{\beta}^{\pi}V \right] \left( s \right) \label{eq:Bellamn optimality operator}
\end{align*}
\begin{lemma} Given any value function $V:\mathcal{S} \rightarrow \mathbb{R}$,
\begin{align}
\forall s\in \mathcal{S} : [\mathcal{T} _{\beta}V]\left( s \right) =\beta \log \mathbb{E} _{a\sim \pi _{\mathrm{ref}}\left( \cdot |s \right)}\left[ \exp \left\{ \frac{1}{\beta}\left( r\left( s,a \right) +V\left( f(s,a) \right) \right) \right\} \right]
\numberthis
\end{align}
\label{lemma:Bellamn optimality operator}
\end{lemma}

\paragraph{Performance difference lemma} Following is the key theoretical analysis tool used in this work,

\begin{lemma} For arbitrary two policies $\pi, \tilde{\pi} \in \Pi$ and any distribution  $\rho \in \Delta(\mathcal{S})$,
\begin{align}
V_{\beta}^{\pi}\left( \rho \right) -V_{\beta}^{\tilde{\pi}}\left( \rho \right) =\mathbb{E} _{\rho}^{\pi}\left[ \mathcal{T} _{\beta}^{\pi}V_{\beta}^{\tilde{\pi}}\left( s \right) -V_{\beta}^{\tilde{\pi}}\left( s \right) \right].
\label{eq:pdl}
\end{align}
\label{lemma:pdl}
\end{lemma}
The proof of this lemma is deferred to the Appendix.

\paragraph{Optimal policy} In the following we give the optimality conditions of the step-level RLHF.
\begin{theorem}[\cite{Agarwal_pg_2019,pg-liu}]
if a policy $\hat{\pi} \in \Pi$ satisfies
\begin{align*}
\forall s\in \mathcal{S} :V_{\beta}^{\hat{\pi}}\left( s \right) =[\mathcal{T} _{\beta}V_{\beta}^{\hat{\pi}}]\left( s \right),
\end{align*}
then $\hat{\pi}=\pi_\beta^*$. Besides, the optimal policy $\pi_\beta^*$ has the unique analytic form of 
\begin{align*}
\forall s\in \mathcal{S} ,a\in \mathcal{A} : \log \frac{\pi _{\beta}^{*}\left( a|s \right)}{\pi _{\mathrm{ref}}\left( a|s \right)}=\frac{1}{\beta}\left( Q^*_\beta\left( s,a \right) -V_{\beta}^{*}\left( s \right) \right).
\end{align*}
\label{theorem:optimal policy}
\end{theorem}

\section{Direct Advantage Policy Optimization}
\label{Direct Advantage Optimization}
In this section, we present the details of DAPO. We only consider the case that the groud-truth reward is only assigned in the terminal state since it's the most common situation in the LLMs setting. We first derive the objective of policy optimization in Section \ref{subsection: Policy Objective} assuming that one has access to the true value function. Then in Section \ref{subsection: Critic Objective}, we present how to train a  critic as the approximation of the true value function. Finally, we give our theoretical analysis of DAPO in Section \ref{subsection: theoretical analysis}. 

\subsection{Policy Objective}
\label{subsection: Policy Objective}

We first derive the policy optimization objective in one single iteration provided with the reference policy $\pi _{\mathrm{ref}}$. In the perspective of policy optimization, the most important quantity may be the advantage function. Recall Lemma \ref{lemma:pdl}. The performance difference for any policy $\pi \in \Pi$ w.r.t $\pi _{\mathrm{ref}}$ is measured as
\begin{align*}
V_{\beta}^{\pi}\left( \mu \right) -V_{\beta}^{\pi _{\mathrm{ref}}}\left( \mu \right) :=\mathbb{E} _{\mu}^{\pi}\left[ \mathcal{I} _s\left( \pi ,\pi _{\mathrm{ref}} \right) \right],
\end{align*}
where the term $\mathcal{I} _s\left( \pi ,\pi _{\mathrm{ref}} \right) :=\mathcal{T} _{\beta}^{\pi}V_{\beta}^{\pi _{\mathrm{ref}}}\left( s \right) -V_{\beta}^{\pi _{\mathrm{ref}}}\left( s \right)$ can be regarded as the \textbf{one-step policy improvement} at the state $s$. A direct computation yields that
\begin{align*}
\mathcal{I} _s\left( \pi ,\pi _{\mathrm{ref}} \right) &=\mathbb{E} _{a\sim \pi \left( \cdot |s \right)}\left[ r\left( s,a \right) +V_{\beta}^{\pi _{\mathrm{ref}}}\left( f\left( s,a \right) \right) -\beta \log \frac{\pi \left( a|s \right)}{\pi _{\mathrm{ref}}\left( a|s \right)} \right] -V_{\beta}^{\pi _{\mathrm{ref}}}\left( s \right) 
\\
\,\,            &\overset{\left( a \right)}{=}\mathbb{E} _{a\sim \pi \left( \cdot |s \right)}\left[ A_{\beta}^{\pi _{\mathrm{ref}}}\left( s,a \right) -\beta \log \frac{\pi \left( a|s \right)}{\pi _{\mathrm{ref}}\left( a|s \right)} \right] 
\\
\,\,            &\overset{\left( b \right)}{=}\mathbb{E} _{a\sim \pi \left( \cdot |s \right)}\left[ A^{\pi _{\mathrm{ref}}}\left( s,a \right) -\beta \log \frac{\pi \left( a|s \right)}{\pi _{\mathrm{ref}}\left( a|s \right)} \right]
\numberthis
\label{def: one-step policy improvement}
\end{align*}
where (a),(b) leverages the definition of $A_{\beta}^{\pi _{\mathrm{ref}}}$ and $A^{\pi _{\mathrm{ref}}}$ in \eqref{def:advantage_function}. A natural idea that ensures $
V_{\beta}^{\pi}\left( \mu \right) - V_{\beta}^{\pi _{\mathrm{ref}}}\left( \mu \right) \ge 0$ might be to increase $\mathcal{I} _s\left( \pi ,\pi _{\mathrm{ref}} \right)$ on as many states $s$ as possible. For ease of understanding, we first focus on how to increase $\mathcal{I} _s\left( \pi ,\pi _{\mathrm{ref}}\right)$ at an arbitrary state $s \in \mathcal{S}$. Considering parameterized policy $\pi_\theta$, 
the most directly and easily method may be to do multi gradient ascent steps w.r.t $\mathcal{I} _s\left( \pi_\theta ,\pi _{\mathrm{ref}}\right)$, i.e.,
\begin{align}
\forall k\in \mathbb{N} ^+:\theta _{k+1}\gets \theta _k+\eta \cdot \nabla _{\theta}\mathcal{I} _s\left( \pi _{\theta _k},\pi _{\mathrm{ref}} \right).
\numberthis
\label{eq:gradient ascent w.r.t one-step improvement}
\end{align}
\and the parameter after $K\in \mathbb{N}^+$ gradient ascent steps is actually  
\begin{align}
\theta _K=\theta _0-\eta\sum_{k=0}^{K-1}{\nabla _{\theta}\mathcal{I} _{\theta}\left( \pi _{\theta _k},\pi _{\mathrm{ref}} \right) .}
\label{eq: multiple-one-step-policy-improvement}
\end{align}
Following lemma gives an equivalent form of $ \nabla _{\theta}\mathcal{I} _s\left( \pi _{\theta _k},\pi _{\mathrm{ref}} \right)$.
\begin{lemma} If we define the surrogate function
\begin{align*}
\mathcal{H} _{s}^{k}\left( \pi _{\theta},\pi _{\mathrm{ref}} \right) :=\frac{1}{2}\mathbb{E} _{a\sim \pi _{\theta _k}\left( \cdot |s \right)}\left[ \left( \frac{1}{\beta}A^{\pi _{\mathrm{ref}}}\left( s,a \right) -\log \frac{\pi _{\theta}\left( a|s \right)}{\pi _{\mathrm{ref}}\left( a|s \right)} \right) ^2 \right],
\end{align*}    
then $\nabla _{\theta}\mathcal{I} _s\left( \pi _{\theta _k},\pi _{\mathrm{ref}} \right) =-\beta\nabla _{\theta}\mathcal{H}^k_s\left( \pi _{\theta},\pi _{\mathrm{ref}} \right).$
\label{lemma : equivalent_update}
\end{lemma}
By Lemma \ref{lemma : equivalent_update}, the update \eqref{eq: multiple-one-step-policy-improvement} is equivalent to 
\begin{align}
\theta _K=\theta _0-\eta \beta\sum_{k=0}^{K-1}{\nabla _{\theta}\mathcal{H}^k_s\left( \pi _{\theta _k},\pi _{\mathrm{ref}} \right)}.
\label{eq : equivalent_update 2}
\end{align}
We now to turn to the offline policy optimization problem setting for the sake of implementation simplicity and data efficiency. A natural question is that can we implement \eqref{eq : equivalent_update 2} approximately in an offline dataset without sampling the on-policy rollouts at each iteration? Our answer is affirmative. 
We find that (which will be proved latter), just modifying the distribution in the expectation of $\mathcal{H}^k_s$ to a fixed exploratory sampling distribution $\nu(\cdot|s) \in \Delta(\mathcal{A})$ and solving the problem
\begin{align}
\underset{\theta}{\min}\,\,\left\{ \mathcal{L} _s\left( \theta \right) :=\frac{1}{2}\mathbb{E} _{a\sim \nu \left( \cdot |s \right)}\left[ \left( \frac{1}{\beta}A^{\pi _{\mathrm{ref}}}\left( s,a \right) -\log \frac{\pi _{\theta}\left( a|s \right)}{\pi _{\mathrm{ref}}\left( a|s \right)} \right) ^2 \right] \right\},
\label{eq: step-dapo-single state}
\end{align}
can still yield a policy that $\mathcal{I} _s\left( \pi_\theta ,\pi _{\mathrm{ref}} \right) > 0$ unless $\pi_{\text{ref}}=\pi^*_\beta$. Another simple intuition behind \eqref{eq: step-dapo-single state} is that: In order to minimize $\mathcal{L} _s\left( \theta \right)$, for any action $a$ that $\nu(a|s)>0$, one should increase $\log \frac{\pi _{\theta}\left( a|s \right)}{\pi _{\mathrm{ref}}\left( a|s \right)}$ if $
A^{\pi _{\mathrm{ref}}}\left( s,a \right) > 0$ and decrease $\log \frac{\pi _{\theta}\left( a|s \right)}{\pi _{\mathrm{ref}}\left( a|s \right)}$ if $
A^{\pi _{\mathrm{ref}}}\left( s,a \right) < 0$, thus results in the increase of $\mathcal{I} _s\left( \pi_\theta ,\pi _{\mathrm{ref}} \right)$. Meanwhile, it's worthy to note that optimizing \eqref{eq: step-dapo-single state} is often not to try to fit the distribution $
\hat{\pi}\left( \cdot|s \right) =\pi _{\mathrm{ref}}\left( \cdot|s \right) \cdot \exp \left\{ \frac{1}{\beta}A^{\pi _{\mathrm{ref}}}\left( s,\cdot \right) \right\}$ since $\hat{\pi}$
is generally not a valid policy because
$$
\sum_{a\in \mathcal{A}}{\hat{\pi}\left( a|s \right)}=\mathbb{E} _{a\sim \pi _{\mathrm{ref}}\left( \cdot |s \right)}\left[ \exp \left\{ \frac{1}{\beta}A^{\pi _{\mathrm{ref}}}\left( s,a \right) \right\} \right] \ge \exp \left\{ \frac{1}{\beta}\mathbb{E} _{a\sim \pi _{\mathrm{ref}}\left( \cdot |s \right)}\left[ A^{\pi _{\mathrm{ref}}}\left( s,a \right) \right] \right\} =1,
$$
where the last inequality is due to Jensen inequality and the fact that $
\mathbb{E} _{a\sim \pi _{\mathrm{ref}}\left( \cdot |s \right)}\left[ A^{\pi _{\mathrm{ref}}}\left( s,a \right) \right] =0$. Taking the consideration of optimizing multiple states simultaneously and the approximation of advantage function $A^{\pi_{\text{ref}}}$, we finally reach the objective of DAPO:
\begin{align}
\text{DAPO}:\quad
\underset{\theta}{\min}\,\,\mathcal{L} \left( \theta \right) :=\frac{1}{2} \mathbb{E} _{\left( s,a,\hat{A} \right) \sim \mathcal{D}}\left[ \left( \frac{1}{\beta}\hat{A}-\log \frac{\pi _{\theta}\left( a|s \right)}{\pi _{\mathrm{ref}}\left( a|s \right)} \right) ^2 \right] 
\label{eq: step-dapo objective}
\end{align}
Here $\mathcal{D} =\left\{ \left( s_i,a_i,\hat{A}_i \right) \right\} _{i=1}^{N}$ is a pre-collected offline training dataset, $\hat{A}_i$ is an estimation of true advantage $
A^{\pi _{\mathrm{ref}}}(s_i,a_i)$. We present our theoretical analysis of problem \eqref{eq: step-dapo objective} in Section \ref{subsection: theoretical analysis} and the detailed implementation in Section \ref{subsection: Experiments}. Recall the definition of advantage function in \ref{def:advantage_function},
\begin{align*}
A^{\pi _{\mathrm{ref}}}\left( s,a \right) =Q^{\pi _{\mathrm{ref}}}\left( s,a \right) -V^{\pi _{\mathrm{ref}}}\left( s \right) =Q^{\pi _{\mathrm{ref}}}\left( s,a \right) -\mathbb{E} _{a^{\prime}\sim \pi _{\mathrm{ref}}}\left[ Q^{\pi _{\mathrm{ref}}}\left( s,a^{\prime} \right) \right],
\end{align*}
and the definition of Q function in \eqref{def:Q_function}  
\begin{align*}
Q^{\pi _{\mathrm{ref}}}\left( s,a \right) =r\left( s,a \right) +V^{\pi _{\mathrm{ref}}}\left( f\left( s,a \right) \right).
\end{align*}
Notice that for non-terminal states, $r\left( s,a \right) =0$. 
Thus in our implementation, we sample multiple actions $
\left\{ a_i \right\} _{i=1}^{M}$ for each state $s$ from $\pi_{\text{ref}}$ and estimate the advantage of each action by 
\begin{align}
\forall i\in \left[ m \right] :\hat{A}\left( s,a_i \right) :=V_{\phi}\left( f\left( s,a_i \right) \right) -\frac{1}{M}\sum_{j=1}^M{\left( V_{\phi}\left( f\left( s,a_j \right) \right) \right)}.
\label{eq: advantage estimation}
\end{align}
Here $V_\phi \approx V^{\pi_{\text{ref}}}$ is a pretrained critic function. Since \eqref{eq: advantage estimation} is fully determined by the critic $V_\phi$, thus the training of $V_\phi$ plays a crucial role for the optimization of training states. In the following, we introduce the optimization method of $V_\phi$.

\subsection{Critic Objective}
\label{subsection: Critic Objective}

Generally speaking, the critic optimization problem for any target policy $\pi$ can be formulated as 
\begin{align}
\underset{\phi}{\argmin} \,\mathbb{E} _{s\sim \mathcal{D}}\left[ \mathcal{L} \left( V_{\phi}\left( s \right) ,V^{\pi}\left( s \right) \right) \right],
\label{eq: critic optimization general}
\end{align}
here $\mathcal{L}$ is a sample-wise loss function, $\mathcal{D}$ is the distribution of training states. However, since the target $V^{\pi_{\text{ref}}}(s)$ can not be accessed to in advance, we need to construct the estimate of $V^{\pi}$ as the training samples for the optimization of $V_\phi$. One can show that
\begin{align*}
V^{\pi}\left( s \right) =\mathbb{E} _{\forall t\in \mathbb{N} :a_t\sim \pi \left( \cdot |s_t \right)}\left[ \sum_{t=0}^{T-1}{r\left( s_t,a_t \right)}|s_0=s \right] =\mathbb{P} ^{\pi}\left( r\left( s_T \right) =1|s_0=s \right).
\end{align*}
Thus, one can firstly use a behavior policy $\pi_{\text{gen}}$ (also called \textbf{generator}) to generate a set of training states $s$. Then for each training state $s$, one can use $\pi$ as the \textbf{completer} to sample $N$ sub-trajectories from $s$, 
\begin{align*}
\left\{ \tau _i=\left( s,a_{0}^{\left( i \right)},a_{1}^{\left( i \right)},...,a_{T_i-1}^{\left( i \right)} \right) \,\,| \forall j\in \left[ T_i \right] ,s_{j}^{\left( i \right)}=f\left( s_{j-1}^{\left( i \right)},a_{j-1}^{\left( i \right)} \right) , s_{0}^{\left( i \right)}=s \right\} _{i=1}^{N},
\end{align*}
and construct the empirical MC mean,
\begin{align*}
\mathrm{MC}_N\left( s \right) :=\frac{1}{N}\sum_{i=1}^N{\mathbf{1}\left\{ s_{T_i}^{\left( i \right)}=1 \right\}},
\end{align*}
as the estimation of $V^\pi(s)$, which is guaranteed to converge to $V^\pi(s)$ as $n \rightarrow \infty$  by the strong law of large numbers. Notice that the state value $V^\pi(s)$ is a probability. Thus we the binary cross-entropy loss, i.e.,
\begin{align*}
\mathcal{L} _{\mathrm{BCE}}\left( y,y_{\mathrm{pred}} \right) =-\left( y\log \left( y_{\mathrm{pred}} \right) +\left( 1-y \right) \log \left( 1-y_{\mathrm{pred}} \right) \right) 
\end{align*}
as the sample-wise loss function $\mathcal{L}$ in \eqref{eq: critic optimization general}. Finally, the critic optimization objective is 
\begin{align*}
\underset{\phi}{\min}\,\,\mathbb{E} _{s\sim \mathcal{D}}\left[ \mathcal{L} _{\mathrm{BCE}}\left( \mathrm{MC}_N\left( s \right) ,V_{\phi}\left( s \right) \right) \right].
\end{align*}
\begin{remark}[Should we treat $V_\phi$ as a PRM?] Its worthy to note that the critic training method we presented above is mostly aligned with \cite{Math-Shepherd}. However, we argue that the $V_\phi$ should be treated as a value function other than a process reward model as in \cite{Math-Shepherd}. Here are the reasons : 1) From the optimization perspective, $V_\phi$ is trained to approximate the true state value function of the completer. 2) From the RL perspective, using $V_\phi$ as a reward function will deviate from the original RLHF optimization objective and easily lead to a serve reward hacking issue on the number of reasoning steps as shown in \cite{prm_PPO}.

\end{remark}

\subsection{Theoretical Analysis}
\label{subsection: theoretical analysis}
In this section, we give some theoretical analysis of DAPO in exact setting and the optimization problem in \eqref{eq: step-dapo objective} becomes
\begin{align}
\text{Exact DAPO}:\quad
\underset{\pi \in \Pi}{\min}\,\,\frac{1}{2}\cdot \mathbb{E} _{s\sim \nu _{\mathcal{S}}}\mathbb{E} _{a\sim \nu _{\mathcal{A}}\left( \cdot |s \right)}\left[ \left( \frac{1}{\beta}A^{\pi _{\mathrm{ref}}}\left( s,a \right) -\log \frac{\pi \left( a|s \right)}{\pi _{\mathrm{ref}}\left( a|s \right)} \right) ^2 \right].
\label{eq: exact-step-dapo objective}
\end{align}
Here $\nu_\mathcal{S}$ is the distribution of training states, $\nu_\mathcal{A}(\cdot|s)$ the distribution of training actions at each state $s$. We assume both $\nu _{\mathcal{S}}$ and $\nu _{\mathcal{A}}\left(\cdot|s \right)$ are exploratory, i.e., 
$$\forall s\in \mathcal{S} ,a\in \mathcal{A} : \nu _{\mathcal{S}}\left( s \right) >0, \nu _{\mathcal{A}}\left( a|s \right) >0.$$
Notice that the state space $\mathcal{S}$ and action space $\mathcal{A}$ are finite in step-level MDP. Thus the assumption that the training distribution is exploratory isn't strict in exact setting. For example, the assumption can be satisfied if  $\nu_\mathcal{A} $ is a softmax policy and $\nu_\mathcal{S}$ is the state visitation distribution induced by $\nu_\mathcal{A}$. We first present  monotonic improvement property when one has access to the advantage function oracle $A^{\pi _{\mathrm{ref}}}$.

\begin{theorem}[Monotonic improvement] Suppose $\forall s\in \mathcal{S}, a \in \mathcal{A},\pi_{\text{ref}}(a|s) > 0$. Let $\pi ^+$ be the solution in \eqref{eq: exact-step-dapo objective}. Then for any state $s \in \mathcal{S}$, there exists a function $
\lambda _s: \Delta \left( \mathcal{A} \right) \rightarrow \left[ 0,+\infty \right)$ such that 
\begin{align*}
V_{\beta}^{\pi ^+}\left( \mu \right) -V_{\beta}^{\pi _{\mathrm{ref}}}\left( \mu \right) \ge \mathbb{E} _{\mu}^{\pi ^+}\left[ \lambda _s\left( \nu _{\mathcal{A}} \right) \right] \ge 0.
\end{align*}
The equality holds if and only if $\pi_{ref}=\pi^*_\beta$.
\label{theorem:monotonic improvement}
\end{theorem}

\begin{remark} The function $\lambda_s$  serves as a lower bound of the one step policy improvement $
\mathcal{I} _s\left( \pi ^+,\pi _{\mathrm{ref}} \right) $ in each states $s$. Since Theorem \ref{theorem:monotonic improvement} is established for arbitrary exploratory $\nu_\mathcal{A}$, $\lambda_s$ should be a function of $\nu_\mathcal{A}$. For the case  $\nu _{\mathcal{A}}\left( \cdot |s \right) =\pi _{\mathrm{ref}}\left( \cdot |s \right)$, one can show that 
\begin{align*}
\lambda _s\left(\pi_{\text{ref}} \right) =\beta \cdot \mathrm{KL}\left( \pi _{\mathrm{ref}}\left( \cdot |s \right) ,\pi ^+\left( \cdot |s \right) \right) 
\end{align*}
by combining \eqref{eq:dual eq} and \eqref{eq:1} in the proof of Theorem \ref{theorem:monotonic improvement}. It remains as a great interest for us to derive a more precise expression of $\lambda_s$ for general $\nu_\mathcal{A}$ and figure out which $\nu_\mathcal{A}$ should yield biggest lower bound.
\end{remark}
Theorem \ref{theorem:monotonic improvement} actually implies that once the training distribution is exploratory, DAPO will improve the performance until there is no room for improvement in the trust region of $\pi_{\text{ref}}$. Theorem \ref{theorem:monotonic improvement} also establishes the monotonic improvement property on unregularized state value function. This can be verified by
\begin{align*}
V^{\pi ^+}\left( \mu \right) &=V_{\beta}^{\pi ^+}\left( \mu \right) +\mathbb{E} _{\mu}^{\pi ^+}\left[ \mathrm{KL}\left( \pi ^+\left( \cdot |s \right) ,\pi _{\mathrm{ref}}\left( \cdot |s \right) \right) \right] \ge V_{\beta}^{\pi ^+}\left( \mu \right) \ge V_{\beta}^{\pi _{\mathrm{ref}}}\left( \mu \right) =V^{\pi _{\mathrm{ref}}}\left( \mu \right).
\end{align*}

\section{Experiments}
In this section, we conduct DAPO on mathematical and coding datasets individual. The experimental setup will be detailed in Section \ref{subsection : Experimental setup}, while the specific conclusions and benchmark results of our experiments will be presented in Section \ref{subsection : exp results}.

\subsection{Experimental Setup}
\label{subsection : Experimental setup}

\paragraph{Training Dataset} For all the DAPO experiments on mathematics, we utilize only the 7500 training problems from the dataset MATH \cite{DATASET_MATH} to generate the advantage datasets for DAPO training, requiring no additional human annotations. In particular, we only use the questions and the corresponding  golden answers in the dataset while the provided solutions are not used for training. For coding experiments, For coding experiments, we curate the TACO \cite{li2023taco} dataset to compile approximately 4,000 competition-level programming questions derived from real-world scenarios. We utilize its provided unit test cases to evaluate whether a solution is accurate.

\paragraph{Base Models} Considering the reproducibility and of DAPO, our experiments are taken over of several open-source language models including general and math-specific models. For general models, we consider META-LLama-3.1-8B-Instruct \cite{LLAMA} , OpenO1-LLama-8B-v0.1 \cite{OPEN-O1}, Skywork-O1-Open-LLama3.1-8B \cite{skyworkopeno12024}, Qwen2.5-72B-Instruct \cite{qwen2.5}. For math-specific models, we consider Skywork-Math-LLama \cite{skywork_math}, Qwen2-Math-7B-Instruct \cite{qwen2} and Qwen2.5-Math-7B-Instruct \cite{Qwen2.5-math}. Notice that Skywork-O1-Open-LLama3.1-8B, Qwen2-Math-7B-Instruct and Qwen2.5-Math-7B-Instruct are already trained by RL algorithms \cite{DeepSeekMath}. Thus we conduct continue-RL training on three models by DAPO.

\paragraph{Benchmarks \& Metrics} For mathematical experiments, We evaluate our training performance on English  mathematical benchmarks. In addition to the 5000 test problems from MATH \cite{DATASET_MATH}, We also add 4 out-of-domain benchmarks, GSM8K \cite{DATASET_GSM8K}, Minerva Math \cite{DATASET_Minerva_Math}, Olympiad Bench \cite{DATASET_olympid_bench} and College Math \cite{DATASET_colledge_math}, to test the performance generalization of DAPO. For coding experiments, we evaluate DAPO on several widely-used benchmarks in
code generation, i.e., HumanEval \cite{BENCHMARK_HUMANEVAL}, HumanEval+ \cite{BENCHMARK_HUMANEVAL_PLUS}, MBPP \cite{BENCHMARK_MBPP}, MBPP+ \cite{BENCHMARK_MBPP_PLUS} and LiveCodeBench \cite{BENCHMARK_LIVECODE}. All the evaluations are conducted in a zero-shot greedy sampling (i.e. temperature 0) setting with a cot prompt template and a maximum amount of 2048 newly generated tokens during evaluation (except 4096 for OpenO1-LLama-8B-v0.1 and Skywork-O1-Open-LLama3.1-8B).

\begin{figure}
    \centering
    \includegraphics[width=0.98\linewidth]{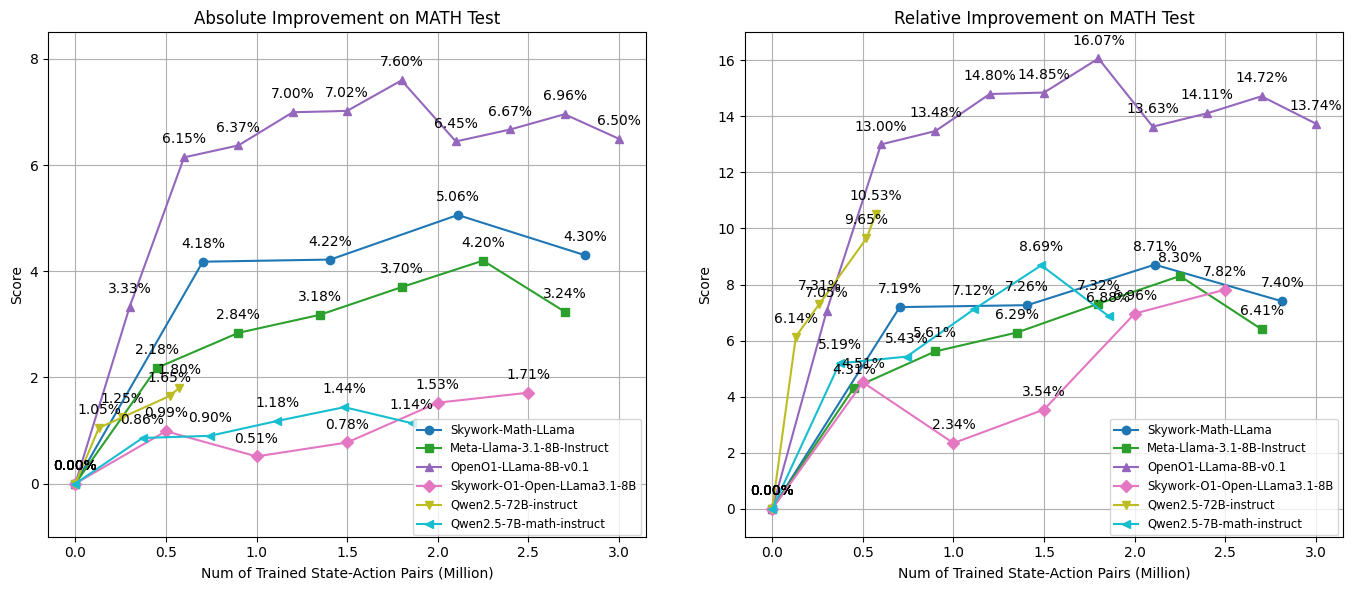}
    \caption{\textbf{Performance improvement on MATH TEST during training process.} Let $x$ be the accuracy of base model on MATH TEST and $y$ be the accuracy after DAPO training. Absolute improvement refers to $y-x$ and relative improvement refers to $\frac{y-x}{1-x}$.} 
    \label{fig:performance improvement during training}
\end{figure}

\paragraph{DAPO Implementations} For each DAPO experiment, we first use the base model to generate 128 solutions for each training problem and then 6 solutions are selected out of them 
while making the correct solutions and wrong solutions as balanced as possible. For each selected solution, we use one line break  (i.e., "$\backslash\mathrm{n}$") to segment the solution steps and use the base model as the completer to do 16 completions for each reasoning step. Then the MC estimations are constructed for the critic training. The training methodology of critic is presented in Section \ref{subsection: Critic Objective}. In our implementation, the critics are fine-tuned for one epoch on Qwen2.5-Math-7B-Instruct for mathematical experiments and Qwen2.5-coder-7B-Instruct for coding experiments, with a learning rate of 5e-6 and batch size 512. After that, we extract the next steps (action) for each intermediate step (state) in the completion datasets and compute the advantage using the critic for DAPO training. For all experiments, we use the global batch size 2048, learning rate 5e-7 and $\beta=0.01$ or $0.02$. Following are some useful tricks applied in all of the DAPO training.
\begin{itemize}

    \item \textbf{State-wise learning.} Notice that for each training state (intermediate step) $s$, there are multiple actions (next step) for DAPO training. Rather than shuffling all state-action pairs, we try to place different actions from the same state within the same batch for DAPO training in order to provide contrastive gradients.
    
    \item \textbf{Learning uniformly on approximate action space.} Suppose we have $n$ actions $\left\{ a_i \right\} _{i=1}^{n}$ for each training state $s$. Although the true action space $\mathcal{A}$ is extremely large for LLM, we consider the unique actions of $\left\{ a_i \right\} _{i=1}^{n}$ as the approximate action space denoted as $\Tilde{A}_s$. After computing the advantage for each action, we only add the unique actions to the advantage dataset for DAPO training, which means the training action distribution is $\nu_\mathcal{A}(\cdot|s)=\text{Unif}(\Tilde{A}_s)$ in our implementation.

    \item \textbf{Learning on the states with large advantage gap.}  Considering that the approximation error between the prediction of critic network and the true state value, we only learn those states that have a large advantage gap to make sure the actions with higher advantage are actually better than those actions with a lower advantage. We define the advantage gap as 
    \begin{align*}
    \Delta _s:=\underset{a_i\in \tilde{A}_s}{\max}\,\,A\left( s,a_i \right) -\underset{a_j\in \tilde{A}_s}{\min}A\left( s,a_j \right).
    \end{align*}
    In our implementation, we only learn those states with $\Delta_s >= 0.1$
\end{itemize}

\subsection{Main Results}
\label{subsection : exp results}
In this section, we present our main experiment results. The benchmark results are shown in Table \ref{table:math results}, Table \ref{table:code results} and Table \ref{table:iterative DAPO}.

\paragraph{Math} Our results, presented in Table \ref{table:math results} and Table \ref{table:iterative DAPO}, demonstrate that DAPO consistently enhances the performance of the base model across all tested models, including the state-of-the-art 7B model, Qwen2.5-Math-7B-Instruct, on the in-domain benchmark MATH \cite{DATASET_MATH}. Through DAPO training, Skywork-Math-LLama, META-LLama-3.1-8B-Instruct, OpenO1-LLama-8B-v0.1, and Qwen2.5-72B-instruc achieve 50.54\%, 53.62\%, 60.27\%, 84.55\%  greedy decoding accuracy respectively. For RL models, Qwen2-Math-7B-Instruct,Skywork-O1-Open-LLama3.1-8B and  Qwen2.5-Math-7B-Instruct achieve 76.40\%, 79.81\%,and 84.86\% greedy decoding accuracy respectively. Since these models are already trained using RL, our results indicate that DAPO can further enhance its performance, highlighting the effectiveness of DAPO. We also report the performance improvement during the training process on Figure \ref{fig:performance improvement during training}.

\begin{table}[t]
\centering
\resizebox{\textwidth}{!}{%
\begin{tabular}{@{\extracolsep{4pt}}c|c|ccccccc}
\toprule   
\textbf{Model}  & {In Domain} & \multicolumn{4}{c}{Out of Domain} 
\\
 {}  & MATH & GSM8K & Minerva MATH & Olympiad Bench & College Math 
 \\
\midrule
\multicolumn{6}{c}{\textbf{SFT Models}}
 \\ 
\midrule
{ Skywork-Math-LLama}  & 41.90    & 61.49 & 5.51 & 18.68 & 24.85 \\
{ + DAPO}        & 46.88$ ^{\textcolor{IMPROVE}{+4.98}}$  & 67.55$^{\textcolor{IMPROVE}{+6.06}}$ & 7.34$^{\textcolor{IMPROVE}{+1.82}}$ & 22.38 $^{\textcolor{IMPROVE}{+3.70}}$ & 25.87 $^{\textcolor{IMPROVE}{+1.03}}$ 
\\
\midrule
{META-LLama-3.1-8B-Instruct}  & 49.42    & 85.29 &  26.48 & 16.14 & 30.91 \\

{ + DAPO}        & 53.62$^{\textcolor{IMPROVE}{+4.20}}$  & 86.74$^{\textcolor{IMPROVE}{+1.45}}$ & 23.54$^{\textcolor{Red}{-2.94}}$ & 20.01 $^{\textcolor{IMPROVE}{+3.88}}$ & 30.91 $^{\textcolor{IMPROVE}{+1.63}}$ 

\\
\midrule
{OpenO1-LLama-8B-v0.1}    & 52.73 & 85.99 & 29.04 & 19.86  & 29.10 \\ 
 { + DAPO}   & 60.33 $^{\textcolor{IMPROVE}{+7.60}}$ & 88.77 $^{\textcolor{IMPROVE}{+2.78}}$ & 29.42 $^{\textcolor{IMPROVE}{+0.39}}$ & 24.44 $^{\textcolor{IMPROVE}{+4.59}}$ &  32.12 $^{\textcolor{IMPROVE}{+3.02}}$  
 \\
  \midrule
{Qwen2.5-72B-instruct}    & 82.90 & 95.40 & 46.30 & 45.45  & 43.00 \\ 
 { + DAPO}   & 84.70 $^{\textcolor{IMPROVE}{+1.80}}$ & 95.70 $^{\textcolor{IMPROVE}{+0.30}}$ & 50.00 $^{\textcolor{IMPROVE}{+3.70}}$ & 47.70 $^{\textcolor{IMPROVE}{+2.25}}$ &  43.40 $^{\textcolor{IMPROVE}{+0.40}}$ 
 \\
 \midrule
\multicolumn{6}{c}{\textbf{RL Models}}
 \\
\midrule 
 {Qwen2-Math-7B-Instruct}  & 74.46  & 89.38 & 40.07 & 34.36 & 41.59
 \\
 { + DAPO}        & 75.41 $^{\textcolor{IMPROVE}{+0.95}}$ & 89.45$^{\textcolor{IMPROVE}{+0.07}}$ & 37.51$^{\textcolor{Red}{-2.56}}$ & 37.03 $^{\textcolor{IMPROVE}{+2.67}}$ & 42.23 $^{\textcolor{IMPROVE}{+0.64}}$ 
\\
 \midrule
{Skywork-O1-Open-LLama3.1-8B}    & 78.10 & 91.64 & 26.10 & 43.11  & 40.40 \\ 
 { + DAPO}   & 79.81 $^{\textcolor{IMPROVE}{+1.71}}$ & 92.81 $^{\textcolor{IMPROVE}{+1.17}}$ & 29.77 $^{\textcolor{IMPROVE}{+3.67}}$ & 44.76 $^{\textcolor{IMPROVE}{+1.65}}$ &  40.26 $^{\textcolor{red}{-0.14}}$
\\
\midrule 
 {Qwen2.5-Math-7B-Instruct}  &  83.42  & 95.78 & 40.06 & 38.96 &  42.65 \\

{ + DAPO}        & 84.86 $^{\textcolor{IMPROVE}{+1.44}}$ & 96.14$^{\textcolor{IMPROVE}{+0.36}}$ & 41.56$^{\textcolor{IMPROVE}{+1.50}}$ & 41.25 $^{\textcolor{IMPROVE}{+2.29}}$ & 41.97 $^{\textcolor{Red}{-0.67}}$ 
\\
\bottomrule
\end{tabular}
}
\caption{\textbf{Mathematical benchmark results of DAPO} using zero-shot greedy inference.} 
\label{table:math results}
\end{table}

\paragraph{Code generation}  Our results for the code generation task are shown in Table \ref{table:code results}. DAPO improves the coding performance of both META-LLama-3.1-8B-Instruct and OpenO1-LLama-8B-v0.1 on multiple wildly used benchmarks.  From the table, we can see that DAPO has significantly enhanced the performance of the base model on multiple benchmarks. For META-LLama-3.1-8B-Instruct, DAPO has increased by 3.0\% and 2.4\% respectively on HumanEval \cite{BENCHMARK_HUMANEVAL} and HumanEval+ \cite{BENCHMARK_HUMANEVAL_PLUS}, by 5.0\% and 9.2\% respectively on MBPP \cite{BENCHMARK_MBPP} and MBPP+ \cite{BENCHMARK_MBPP_PLUS}, and by 2.1\% on LiveCodeBench \cite{BENCHMARK_LIVECODE}. For OpenO1-LLama-8B-v0.1, DAPO has increased by 2.5\% and 3.6\% respectively on HumanEval \cite{BENCHMARK_HUMANEVAL} and HumanEval+ \cite{BENCHMARK_HUMANEVAL_PLUS}, by 6.1\% and 5.0\% respectively on MBPP \cite{BENCHMARK_MBPP} and MBPP+ \cite{BENCHMARK_MBPP_PLUS}, and decreased by 2.4\% on LiveCodeBench \cite{BENCHMARK_LIVECODE}. Overall, these results strongly demonstrate the effectiveness of DAPO. 

\begin{table}[htbp]
\centering
\resizebox{\textwidth}{!}{%
\begin{tabular}{@{\extracolsep{4pt}}c|cccccccc}
\toprule   
\textbf{SFT Model}  & \multicolumn{5}{c}{Out of Domain} 
\\
 {}  & HumanEval & HumanEval+ & MBPP & MBPP+ & LiveCodeBench
\\
\midrule
{META-LLama-3.1-8B-Instruct}  & 72.0    & 66.5 &  72.0 & 56.9 & 18.8 \\

{ + DAPO}        & 75.0$^{\textcolor{IMPROVE}{+3.0}}$  & 68.9$^{\textcolor{IMPROVE}{+2.4}}$ & 77.0$^{\textcolor{IMPROVE}{+5.0}}$ & 66.1 $^{\textcolor{IMPROVE}{+9.2}}$ & 20.9 $^{\textcolor{IMPROVE}{+2.1}}$ 

\\
\midrule

{OpenO1-LLama-8B-v0.1}    & 69.5 & 61.0 & 69.8 & 58.7  & {16.1} \\ 
 { + DAPO}   & 72.0 $^{\textcolor{IMPROVE}{+2.5}}$ & 64.6 $^{\textcolor{IMPROVE}{+3.6}}$ & 75.9 $^{\textcolor{IMPROVE}{+6.1}}$ & 63.8 $^{\textcolor{IMPROVE}{+5.1}}$ &  {13.7} $^{\textcolor{Red}{-2.4}}$
 \\
\bottomrule
\end{tabular}
}

\caption{\textbf{Coding benchmark results of DAPO} using zero-shot greedy inference.} 
\label{table:code results}
\end{table}

\paragraph{On the performance generalization of DAPO.} Notice
that the training problems we used for mathematical experiments are only 7500 problems from MATH \cite{DATASET_MATH}. However, DAPO still improves the mathematical performance for base models (especially SFT models) in various benchmarks, exhibiting the strong performance generalization of DAPO. DAPO also improves the  performance of META-LLama-3.1-8B-Instruct and OpenO1-LLama-8B-v0.1 on multiple coding benchmarks simultaneously. It's reasonable to anticipate a higher performance improvement of DAPO with a larger training query set.

\paragraph{On the iterative DAPO.} 
 As suggested by our theoretical results (Theorem \ref{theorem:monotonic improvement}), there still remains the room for performance improvement if $\pi_{\text{ref}}$ is not the optimal policy. Thus we conduct two iterative DAPO experiments on  Skywork-Math-LLama and Qwen2-Math-7B-Instruct.
 At iteration 2, we use the model outputted by DAPO at iteration 1 as $\pi_{\text{ref}}$. Our experiment results are
 summarized in Table \ref{table:iterative DAPO}. We can conclude from Table \ref{table:iterative DAPO} that DAPO can further improves the performance of Skywork-Math-LLama and Qwen2-Math-7B-Instruct. However, due to the limitation of resources, we do not conduct iterative DAPO experiments on more base models and more iterations.

\begin{table}[htbp]
\centering
\resizebox{\textwidth}{!}{%
\begin{tabular}{@{\extracolsep{4pt}}c|c|ccccccc}
\toprule   
\textbf{Model}  & {In Domain} & \multicolumn{4}{c}{Out of Domain} 
\\
 {}  & MATH & GSM8K & Minerva MATH & Olympiad Bench & College Math 
 \\ 
\midrule
{ Skywork-Math-LLama}  & 41.90    & 61.49 & 5.51 & 18.68 & 24.85 \\
{ + DAPO iter 1}        & 46.88$ ^{\textcolor{IMPROVE}{+4.98}}$  & 67.55$^{\textcolor{IMPROVE}{+6.06}}$ & 7.34$^{\textcolor{IMPROVE}{+1.82}}$ & 22.38 $^{\textcolor{IMPROVE}{+3.70}}$ & 25.87 $^{\textcolor{IMPROVE}{+1.03}}$ 
\\
{ + DAPO iter 2}        & $50.54 ^{\textcolor{IMPROVE}{+8.64}}$  & $69.04 ^{\textcolor{IMPROVE}{+7.55}}$ & 8.09$^{\textcolor{IMPROVE}{+2.58}}$ & 23.86 $^{\textcolor{IMPROVE}{+5.18}}$ & 27.20 $^{\textcolor{IMPROVE}{+2.35}}$ 
 \\
\midrule 
 {Qwen2-Math-7B-Instruct}  & 74.46  & 89.38 & 40.07 & 34.36 & 41.59
 \\
 { + DAPO iter 1}        & 75.41 $^{\textcolor{IMPROVE}{+0.95}}$ & 89.45$^{\textcolor{IMPROVE}{+0.07}}$ & 37.51$^{\textcolor{Red}{-2.56}}$ & 37.03 $^{\textcolor{IMPROVE}{+2.67}}$ & 42.23 $^{\textcolor{IMPROVE}{+0.64}}$ 
 
 \\
{ + DAPO iter 2}        & 76.40 $^{\textcolor{IMPROVE}{+1.94}}$ & 89.30$^{\textcolor{Red}{-0.08}}$ & 38.25$^{\textcolor{Red}{-1.82}}$ & 37.64 $^{\textcolor{IMPROVE}{+3.27}}$ & 41.16 $^{\textcolor{Red}{-0.43}}$ 
\\
\bottomrule
\end{tabular}
}
\caption{\textbf{Mathematical benchmark results of iterative DAPO} using zero-shot greedy inference.} 
\label{table:iterative DAPO}
\end{table}

\label{subsection: Experiments}
\label{Experiments}

\section{Discussion, Limitations, and Future Work }
\label{Discussion}

In this work, we propose an offline step-level RLHF method called Direct Advantage Policy Optimization (DAPO), aimed at optimizing the generation of reasoning steps. DAPO has achieved significant performance improvements on both mathematical and coding benchmarks, thereby demonstrating its effectiveness. Compared with standard response-level methods such as DPO,DRO, 
DAPO leverages the critic function to achieve more fine-grained policy optimization. Compared with other step-level RLHF methods, DAPO split the training  of actor and critic into two individual stages which stabilize the RL training process and achieve the performance improvement at the same time. DAPO has limitations that also provide avenues for future work. 
First, as we mentioned before, we only use the a little training queries for DAPO training. whether there exists a scaling law of performance improvement on training queries remains as a great interest for us. Second, we did not conduct iterative DAPO on more models or on more iterations due to the limit of computation resources. We would like to investigate what's the limitation of DAPO on a certain number of training problems in future work. Third, the computation cost is still high for the whole procedure of DAPO. There still need find a more computation efficient way to implement DAPO.

\bibliographystyle{plain}
\bibliography{refs}

\appendix
\newpage
\begin{center}
    \Large{\textbf{Appendix}}
\end{center}
\section{Proofs}

\subsection{Proof of Lemma \ref{lemma:Bellamn optimality operator}}
By the definition of $\mathcal{T}_\beta$, for any state $s \in \mathcal{S}$, we define
\begin{align*}
\pi _{\beta}^{V}\left( \cdot |s \right) \in \underset{p\in \Delta \left( \mathcal{A} \right)}{\mathrm{arg}\max}\,\,\mathbb{E} _{a\sim p}\left[ r\left( s,a \right) +V\left( f(s,a) \right) -\beta \log \frac{p\left( a \right)}{\pi _{\mathrm{ref}}\left( a|s \right)} \right].
\end{align*}
One can easily find that 
\begin{align*}
\forall s\in \mathcal{S} ,a\in \mathcal{A} : \pi _{\beta}^{V}\left( a|s \right) \,\,:=\frac{\pi _{\mathrm{ref}}\left( a|s \right) \cdot \exp \left\{ \frac{1}{\beta}\left( r\left( s,a \right) +V(f(s,a)) \right) \right\}}{\mathbb{E} _{a^{\prime}\sim \pi _{\mathrm{ref}}\left( \cdot |s \right)}\left[ \exp \left\{ \frac{1}{\beta}\left( r\left( s,a^{\prime} \right) +V(f(s,a)^{\prime}) \right) \right\} \right]}
\end{align*}
and we omit the proof of it here. Then 
\begin{align*}
\mathcal{T} _{\beta}V\left( s \right) &=\mathcal{T} _{\beta}^{\pi _{\beta}^{V}}V\left( s \right) 
\\
\,\,         &=\,\mathbb{E} _{a\sim \pi _{\beta}^{V}\left( \cdot |s \right)}\left[ r\left( s,a \right) +V\left( f(s,a) \right) -\beta \log \frac{\pi _{\beta}^{V}\left( a|s \right)}{\pi _{\mathrm{ref}}\left( a|s \right)} \right] 
\\
\,\,         &=\,\mathbb{E} _{a\sim \pi _{\beta}^{V}\left( \cdot |s \right)}\left[ r\left( s,a \right) +V\left( f(s,a) \right) -\beta \log \left( \frac{\exp \left\{ \frac{1}{\beta}\left( r\left( s,a \right) +V(f(s,a)) \right) \right\}}{\mathbb{E} _{a^{\prime}\sim \pi _{\mathrm{ref}}\left( \cdot |s \right)}\left[ \exp \left\{ \frac{1}{\beta}\left( r\left( s,a^{\prime} \right) +V(f(s,a)^{\prime}) \right) \right\} \right]} \right) \right] 
\\
\,\,         &=\beta \log \left( \mathbb{E} _{a\sim \pi _{\mathrm{ref}}\left( \cdot |s \right)}\left[ \exp \left\{ \frac{1}{\beta}\left( r\left( s,a \right) +V(f(s,a)) \right) \right\} \right] \right) 
\end{align*}

\subsection{Proof of Lemma \ref{lemma:pdl}}
\begin{proof} Recall the definition of state value functions in \eqref{def:value_function}. 
 Then one has
\begin{align*}
V_{\beta}^{\pi}\left( \rho _0 \right) -V_{\beta}^{\tilde{\pi}}\left( \rho _0 \right) &=\mathbb{E} _{\forall t\in \mathbb{N} :a_t\sim \pi \left( \cdot |s_t \right)}\left[ \sum_{t=0}^{T-1}{\left( r\left( s_t,a_t \right) -\beta \mathrm{KL}\left( \pi \left( \cdot |s_t \right) ,\pi _{\mathrm{ref}}\left( \cdot |s_t \right) \right) \right)}|s_0\sim \rho _0 \right] -V_{\beta}^{\tilde{\pi}}\left( \rho _0 \right) 
\\
\,\,                     &=\mathbb{E} _{\forall t\in \mathbb{N} :a_t\sim \pi \left( \cdot |s_t \right)}\left[ \sum_{t=0}^{T-1}{\left( r\left( s_t,a_t \right) -\beta \mathrm{KL}\left( \pi \left( \cdot |s_t \right) ,\pi _{\mathrm{ref}}\left( \cdot |s_t \right) \right) \right)}|s_0\sim \rho _0 \right] 
\\
\,\,                      &-\mathbb{E} _{\forall t\in \mathbb{N} :a_t\sim \pi \left( \cdot |s_t \right)}\left[ \sum_{t=0}^{T-1}{\left( V_{\beta}^{\tilde{\pi}}\left( s_t \right) -V_{\beta}^{\tilde{\pi}}\left( s_{t+1} \right) \right)}|s_0\sim \rho _0 \right] 
\\
\,\,                     &=\mathbb{E} _{\forall t\in \mathbb{N} :a_t\sim \pi \left( \cdot |s_t \right)}\left[ \sum_{t=0}^{T-1}{\left( r\left( s_t,a_t \right) -\beta \mathrm{KL}\left( \pi \left( \cdot |s_t \right) ,\pi _{\mathrm{ref}}\left( \cdot |s_t \right) \right) +V_{\beta}^{\tilde{\pi}}\left( s_{t+1} \right) -V_{\beta}^{\tilde{\pi}}\left( s_t \right) \right)}|s_0\sim \rho _0 \right] 
\\
\,\,                     &=\mathbb{E} _{\forall t\in \mathbb{N} :a_t\sim \pi \left( \cdot |s_t \right)}\left[ \sum_{t=0}^{T-1}{\left( \mathcal{T} _{\beta}^{\pi}V_{\beta}^{\tilde{\pi}}\left( s_t \right) -V_{\beta}^{\tilde{\pi}}\left( s_t \right) \right)}|s_0\sim \rho _0 \right] 
\\
\,\,                    &=\mathbb{E} _{\rho}^{\pi}\left[ \mathcal{T} _{\beta}^{\pi}V_{\beta}^{\tilde{\pi}}\left( s \right) -V_{\beta}^{\tilde{\pi}}\left( s \right) \right]
\end{align*}    
\end{proof}

\subsection{Proof of Lemma \ref{lemma : equivalent_update}}
\begin{proof}
Recall that 
\begin{align*}
\mathcal{H} _{s}^{k}\left( \pi _{\theta},\pi _{\mathrm{ref}} \right) =\frac{1}{2}\mathbb{E} _{a\sim \pi _{\theta _k}\left( \cdot |s \right)}\left[ \left( \frac{1}{\beta}A^{\pi _{\mathrm{ref}}}\left( s,a \right) -\log \frac{\pi _{\theta}\left( a|s \right)}{\pi _{\mathrm{ref}}\left( a|s \right)} \right) ^2 \right] 
\end{align*}
A direct computation yields that 
\begin{align*}
\nabla _{\theta}\mathcal{H} _{s}^{k}\left( \pi _{\theta},\pi _{\mathrm{ref}} \right) &=\frac{1}{2}\mathbb{E} _{a\sim \pi _{\theta _k}\left( \cdot |s \right)}\left[ \nabla _{\theta}\left( \frac{1}{\beta}A^{\pi _{\mathrm{ref}}}\left( s,a \right) -\log \frac{\pi _{\theta}\left( a|s \right)}{\pi _{\mathrm{ref}}\left( a|s \right)} \right) ^2 \right] 
\\
\,\,                  &=\mathbb{E} _{a\sim \pi _{\theta _k}\left( \cdot |s \right)}\left[ \left( \frac{1}{\beta}A^{\pi _{\mathrm{ref}}}\left( s,a \right) -\log \frac{\pi _{\theta}\left( a|s \right)}{\pi _{\mathrm{ref}}\left( a|s \right)} \right) \nabla _{\theta}\left( \frac{1}{\beta}A^{\pi _{\mathrm{ref}}}\left( s,a \right) -\log \frac{\pi _{\theta}\left( a|s \right)}{\pi _{\mathrm{ref}}\left( a|s \right)} \right) \right] 
\\
\,\,                  &=-\mathbb{E} _{a\sim \pi _{\theta _k}\left( \cdot |s \right)}\left[ \left( \frac{1}{\beta}A^{\pi _{\mathrm{ref}}}\left( s,a \right) -\log \frac{\pi _{\theta}\left( a|s \right)}{\pi _{\mathrm{ref}}\left( a|s \right)} \right) \nabla _{\theta}\log \pi _{\theta}\left( a|s \right) \right].
\end{align*}
By \eqref{def: one-step policy improvement},
\begin{align*}
\mathcal{I} _s\left( \pi _{\theta},\pi _{\mathrm{ref}} \right) =\mathbb{E} _{a\sim \pi _{\theta}\left( \cdot |s \right)}\left[ A^{\pi _{\mathrm{ref}}}\left( s,a \right) -\beta \log \frac{\pi _{\theta}\left( a|s \right)}{\pi _{\mathrm{ref}}\left( a|s \right)} \right].
\end{align*}
Then
\begin{align*}
\nabla _{\theta}\mathcal{I} _s\left( \pi _{\theta},\pi _{\mathrm{ref}} \right) &=\nabla _{\theta}\mathbb{E} _{a\sim \pi _{\theta}\left( \cdot |s \right)}\left[ A^{\pi _{\mathrm{ref}}}\left( s,a \right) -\beta \log \frac{\pi _{\theta}\left( a|s \right)}{\pi _{\mathrm{ref}}\left( a|s \right)} \right] 
\\
\,\,                &=\mathbb{E} _{a\sim \pi _{\theta}\left( \cdot |s \right)}\left[ \nabla _{\theta}\log \pi _{\theta}\left( a|s \right) \left( A^{\pi _{\mathrm{ref}}}\left( s,a \right) -\beta \log \frac{\pi _{\theta}\left( a|s \right)}{\pi _{\mathrm{ref}}\left( a|s \right)} \right) \right] 
\\
\,\,                &+\mathbb{E} _{a\sim \pi _{\theta}\left( \cdot |s \right)}\left[ \nabla _{\theta}\left( A^{\pi _{\mathrm{ref}}}\left( s,a \right) -\beta \log \frac{\pi _{\theta}\left( a|s \right)}{\pi _{\mathrm{ref}}\left( a|s \right)} \right) \right] 
\\
\,\,                &=\beta \mathbb{E} _{a\sim \pi _{\theta}\left( \cdot |s \right)}\left[ \nabla _{\theta}\log \pi _{\theta}\left( a|s \right) \left( \frac{1}{\beta}A^{\pi _{\mathrm{ref}}}\left( s,a \right) -\log \frac{\pi _{\theta}\left( a|s \right)}{\pi _{\mathrm{ref}}\left( a|s \right)} \right) \right] 
\\                  &-\beta \mathbb{E} _{a\sim \pi _{\theta}\left( \cdot |s \right)}\left[ \nabla _{\theta}\log \pi _{\theta}\left( a|s \right) \right] 
\\
\,\,                &\overset{\left( a \right)}{=}\beta \mathbb{E} _{a\sim \pi _{\theta}\left( \cdot |s \right)}\left[ \nabla _{\theta}\log \pi _{\theta}\left( a|s \right) \left( \frac{1}{\beta}A^{\pi _{\mathrm{ref}}}\left( s,a \right) -\log \frac{\pi _{\theta}\left( a|s \right)}{\pi _{\mathrm{ref}}\left( a|s \right)} \right) \right],
\end{align*}
where (a) is due to 
\begin{align*}
\mathbb{E} _{a\sim \pi _{\theta}\left( \cdot |s \right)}\left[ \nabla _{\theta}\log \pi _{\theta}\left( a|s \right) \right] =\nabla _{\theta}\mathbb{E} _{a\sim \pi _{\theta}\left( \cdot |s \right)}\left[ 1 \right] =\nabla _{\theta}\left( 1 \right) =0.
\end{align*}
Thus it's easy to verify that $
\nabla _{\theta}\mathcal{I} _s\left( \pi _{\theta _k},\pi _{\mathrm{ref}} \right) =-\beta \nabla _{\theta}\mathcal{H} _{s}^{k}\left( \pi _{\theta},\pi _{\mathrm{ref}} \right) \,\,$ and the proof is completed.

\end{proof}

\subsection{Proof of Theorem \ref{theorem:monotonic improvement}}

\begin{proof} Notice that $\forall s \in \mathcal{S}, \nu_\mathcal{S}(s) > 0$. Then it's obviously that for any state $s \in \mathcal{S}$,
\begin{align*}
\pi ^+\left( \cdot |s \right) \in \underset{p\in \Delta \left( \mathcal{A} \right)}{\mathrm{arg}\min}\,\,\mathbb{E} _{a\sim \nu _{\mathcal{A}}\left( \cdot |s \right)}\left[ \left( \frac{1}{\beta}A^{\pi _{\mathrm{ref}}}\left( s,a \right) -\log \frac{p\left( a \right)}{\pi _{\mathrm{ref}}\left( a|s \right)} \right) ^2 \right] .
\end{align*}
Since $\forall s\in \mathcal{S}, a \in \mathcal{A}, \pi_{\text{ref}}(a|s) > 0$, one has $\forall s\in \mathcal{S}, a \in \mathcal{A},\pi^+(a|s)>0$ otherwise the loss becomes infinity.   For ease of notation, we define 
$$
u_{s,a}:=\log \frac{p\left( a \right)}{\pi _{\mathrm{ref}}\left( a|s \right)}\qquad \mathrm{and}\qquad A_{s,a}:=\frac{1}{\beta}A^{\pi _{\mathrm{ref}}}\left( s,a \right).
$$
Then $\left\{ u_{s,a}^{+}:=\log \frac{\pi ^+\left( a|s \right)}{\pi _{\mathrm{ref}}\left( a|s \right)} \,|\,a\in \mathcal{A} \right\} $ is the solution of the following optimization problem:
\begin{align*}
\underset{u_s:=\left( u_{s,a} \right) _{a\in \mathcal{A}}\in \mathbb{R} ^{\left| \mathcal{A} \right|}}{\min}&:\quad \mathcal{L} _s\left( u_s \right) :=\frac{1}{2}\cdot \mathbb{E} _{a\sim \nu _{\mathcal{A}}\left( \cdot |s \right)}\left[ \left( A_{s,a}-u_{s,a} \right) ^2 \right] 
\\
\,\,          \,\,\mathrm{s}.\mathrm{t}&:\quad \sum_{a\in \mathcal{A}}{p\left( a \right)}=\mathbb{E} _{a\sim \pi _{\mathrm{ref}}\left( \cdot |s \right)}\left[ \exp \left\{ u_{s,a} \right\} \right] =1.
\end{align*}
Denote the  Lagrangian function as 
\begin{align*}
\hat{\mathcal{L}}_s\left( u_s,\lambda \right) :=\mathcal{L} _s\left( u_s \right) +\lambda \cdot \left( \mathbb{E} _{a\sim \pi _{\mathrm{ref}}\left( \cdot |s \right)}\left[ \exp \left\{ u_{s,a} \right\} \right] -1 \right).
\end{align*}
Since it's a convex problem, by KKT conditions, there exists a $\lambda _{s}^{*}\in \mathbb{R}$ such that $
\left( u_{s}^+,\lambda _{s}^{*} \right) $ is the solution of the equations:
\begin{align}
\begin{cases}
	\nabla _{u_s}\hat{\mathcal{L}}_s\left( u_{s}^+,\lambda _{s}^{*} \right) =0\\
	\mathbb{E} _{a\sim \pi _{\mathrm{ref}}\left( \cdot |s \right)}\left[ \exp \left\{ u_{s,a}^+ \right\} \right] =1\\
\end{cases}
\label{eq:kkt condition}
\end{align}
Notice that
\begin{align*}
\nabla _{u_s}\hat{\mathcal{L}}_s\left( u_s,\lambda \right) &=\nabla _{u_s}\mathbb{E} _{a\sim \nu _{\mathcal{A}}\left( \cdot |s \right)}\left[ \frac{1}{2}\left( A_{s,a}-u_{s,a} \right) ^2+\lambda \cdot \frac{\pi _{\mathrm{ref}}\left( a|s \right)}{\nu \left( a|s \right)}\cdot \exp \left\{ u_{s,a} \right\} \right] 
\\
\,\,\,\,               &=\nu _{\mathcal{A}}\left( \cdot |s \right) \odot \left( u_s-A_s+\lambda \cdot \frac{\pi _{\mathrm{ref}}\left( \cdot |s \right)}{\nu _{\mathcal{A}}\left( \cdot |s \right)}\cdot \exp \left\{ u_s \right\} \right) .
\end{align*}
Plugging it into \eqref{eq:kkt condition} yields that
\begin{align}
 \forall a\in \mathcal{A} : A_{s,a}-u_{s,a}^{+}=\lambda _{s}^{*}\cdot \frac{\pi _{\mathrm{ref}}\left( a|s \right)}{\nu _{\mathcal{A}}\left( a|s \right)}\cdot \exp \left\{ u_{s,a}^{+} \right\} .
\label{eq:dual eq}
\end{align}
A direct computation yields that 
\begin{align*}
\lambda _{s}^{*}&=\lambda _{s}^{*}\cdot \mathbb{E} _{a\sim \pi _{\mathrm{ref}}}\left[ \exp \left\{ u_{s,a}^{+} \right\} \right] 
\\
\,\,  &\overset{\left( a \right)}{=}\mathbb{E} _{a\sim \nu _{\mathcal{A}}\left( \cdot |s \right)}\left[ \lambda _{s}^{*}\cdot \frac{\pi _{\mathrm{ref}}\left( a|s \right)}{\nu \left( a|s \right)}\cdot \exp \left\{ u_{s,a}^{+} \right\} \right] 
\\
\,\,  &=\mathbb{E} _{a\sim \nu _{\mathcal{A}}\left( \cdot |s \right)}\left[ \,\,A_{s,a} - u_{s,a}^{+} \right],
\end{align*}
where (a) is due to \eqref{eq:dual eq}. We now to show that $\lambda^*_s \ge 0$. Notice that 
\begin{align*}
\mathbb{E} _{a\sim \pi _{\mathrm{ref}}\left( \cdot |s \right)}\left[ A_{s,a}-u_{s,a}^{+} \right] &=\mathbb{E} _{a\sim \pi _{\mathrm{ref}}\left( \cdot |s \right)}\left[ \frac{1}{\beta}A^{\pi _{\mathrm{ref}}}\left( a|s \right) -\log \frac{\pi ^+\left( a|s \right)}{\pi _{\mathrm{ref}}\left( a|s \right)} \right] 
\\
\,\,                          &=\mathbb{E} _{a\sim \pi _{\mathrm{ref}}\left( \cdot |s \right)}\left[ \log \frac{\pi _{\mathrm{ref}}\left( a|s \right)}{\pi ^+\left( a|s \right)} \right] 
\\
\,\,                          &=\mathrm{KL}\left( \pi _{\mathrm{ref}}\left( \cdot |s \right) ,\pi ^+\left( \cdot |s \right) \right).
\numberthis
\label{eq:1}
\end{align*}
Thus if  $\lambda^*_s < 0$, one has $\forall a\in \mathcal{A} :A_{s,a}-u_{s,a}^{+}<0$ due to \eqref{eq:dual eq} and  
$$
\mathbb{E} _{a\sim \pi _{\mathrm{ref}}}\left[ A_{s,a}-u_{s,a}^{+} \right] <0,
$$
which contradicts to \eqref{eq:1}. Recall the definition of $\mathcal{I}_s(\pi,\pi_\text{ref})$ in \eqref{def: one-step policy improvement}. Thus 
\begin{align*}
\mathcal{I} _s\left( \pi ^+,\pi _{\mathrm{ref}} \right) &=\mathbb{E} _{a\sim \pi ^+\left( \cdot |s \right)}\left[ A^{\pi _{\mathrm{ref}}}\left( s,a \right) -\beta \log \frac{\pi ^+\left( a|s \right)}{\pi _{\mathrm{ref}}\left( a|s \right)} \right] 
\\
\,\,              &=\beta \cdot \mathbb{E} _{a\sim \pi ^+\left( \cdot |s \right)}\left[ A_{s,a}-u_{s,a}^{+} \right] 
\\
\,\,              &\overset{\left( a \right)}{=}\beta \cdot \mathbb{E} _{a\sim \pi ^+\left( \cdot |s \right)}\left[ \lambda _{s}^{*}\cdot \frac{\pi _{\mathrm{ref}}\left( a|s \right)}{\nu _{\mathcal{A}}\left( a|s \right)}\cdot \exp \left\{ u_{s,a}^{+} \right\} \right] 
\\
\,\,              &=\beta \cdot \lambda _{s}^{*}\cdot \mathbb{E} _{a\sim \pi ^+\left( \cdot |s \right)}\left[ \frac{\pi ^+\left( a|s \right)}{\nu _{\mathcal{A}}\left( a|s \right)} \right] 
\\
\,\,              &=\beta \cdot \lambda _{s}^{*}\cdot \mathbb{E} _{a\sim \nu _{\mathcal{A}}\left( \cdot |s \right)}\left[ \left( \frac{\pi ^+\left( a|s \right)}{\nu _{\mathcal{A}}\left( a|s \right)} \right) ^2 \right] 
\\
\,\,              &=\beta \cdot \lambda _{s}^{*}\cdot \mathbb{E} _{a\sim \nu _{\mathcal{A}}\left( \cdot |s \right)}\left[ \left( \frac{\pi ^+\left( a|s \right)}{\nu _{\mathcal{A}}\left( a|s \right)} \right) ^2 \right] 
\\
\,\,              &\ge \beta \cdot \lambda _{s}^{*}
\\
                  &:= \lambda_s(\nu_\mathcal{A})
\numberthis
\label{eq: lower bound of policy improvement}
\end{align*}
where the last inequality is due to
\begin{align*}
\mathbb{E} _{a\sim \nu _{\mathcal{A}}\left( \cdot |s \right)}\left[ \left( \frac{\pi ^+\left( a|s \right)}{\nu _{\mathcal{A}}\left( a|s \right)} \right) ^2 \right] \ge \left( \mathbb{E} _{a\sim \nu _{\mathcal{A}}\left( \cdot |s \right)}\left[ \frac{\pi ^+\left( a|s \right)}{\nu _{\mathcal{A}}\left( a|s \right)} \right] \right) ^2=\left( \sum_{a\in \mathcal{A}}{\pi ^+\left( a|s \right)} \right) ^2=1.
\end{align*}
Plugging \eqref{eq: lower bound of policy improvement} into Lemma \ref{lemma:pdl} directly yields that 
\begin{align*}
V_{\beta}^{\pi ^+}\left( \mu \right) -V_{\beta}^{\pi _{\mathrm{ref}}}\left( \mu \right) =\mathbb{E} _{\mu}^{\pi ^+}\left[ \mathcal{I} _s\left( \pi ,\pi _{\mathrm{ref}} \right) \right] \ge \mathbb{E} _{\mu}^{\pi ^+}\left[ \lambda _s\left( \nu _{\mathcal{A}} \right) \right] \ge 0.
\end{align*}
We now to show that $
V_{\beta}^{\pi ^+}\left( \mu \right) =V_{\beta}^{\pi _{\mathrm{ref}}}\left( \mu \right)$ holds if and only if $\pi_{\text{ref}}$ is already the optimal policy. Notice that 
\begin{align*}
V_{\beta}^{\pi ^+}\left( \mu \right) =V_{\beta}^{\pi _{\mathrm{ref}}}\left( \mu \right) \,\,\Leftrightarrow \,\,\forall s\in \mathcal{S} :  \lambda _s\left( \nu _{\mathcal{A}} \right) =0 \Leftrightarrow \forall s\in \mathcal{S} ,a\in \mathcal{A} :  u_{s,a}^{+}-A_{s,a} = 0.
\end{align*}
Thus plugging it into \eqref{eq:1} yields that
\begin{align*}
\mathrm{KL}\left( \pi _{\mathrm{ref}}\left( \cdot |s \right) ,\pi ^+\left( \cdot |s \right) \right) =0
\end{align*}
and $\forall s,a \in \mathcal{S}\times\mathcal{A}: A^{\pi _{\mathrm{ref}}}\left( s,a \right) =\beta \cdot \log \frac{\pi ^+\left( a|s \right)}{\pi _{\mathrm{ref}}\left( a|s \right)}=0$.  Hence for any state $s \in \mathcal{S}$,
\begin{align*}
\mathcal{T} _{\beta}V_{\beta}^{\pi _{\mathrm{ref}}}\left( s \right) &=\beta \log \mathbb{E} _{a\sim \pi _{\mathrm{ref}}\left( \cdot |s \right)}\left[ \exp \left\{ \frac{1}{\beta}\left( r\left( s,a \right) +V_{\beta}^{\pi _{\mathrm{ref}}}\left( f\left( s,a \right) \right) \right) \right\} \right] 
\\
\,\,           &=\beta \log \mathbb{E} _{a\sim \pi _{\mathrm{ref}}\left( \cdot |s \right)}\left[ \exp \left\{ \frac{1}{\beta}Q_{\beta}^{\pi _{\mathrm{ref}}}\left( s,a \right) \right\} \right] 
\\
\,\,\,\,          &=\beta \log \mathbb{E} _{a\sim \pi _{\mathrm{ref}}\left( \cdot |s \right)}\left[ \exp \left\{ \frac{1}{\beta}Q^{\pi _{\mathrm{ref}}}\left( s,a \right) \right\} \right] 
\\
\,\,\,\,          &=\beta \log \mathbb{E} _{a\sim \pi _{\mathrm{ref}}\left( \cdot |s \right)}\left[ \exp \left\{ \frac{1}{\beta}A^{\pi _{\mathrm{ref}}}\left( s,a \right) +\frac{1}{\beta}V^{\pi _{\mathrm{ref}}}\left( s \right) \right\} \right] 
\\
\,\,\,\,          &=\beta \log \mathbb{E} _{a\sim \pi _{\mathrm{ref}}\left( \cdot |s \right)}\left[ \exp \left\{ \frac{1}{\beta}V^{\pi _{\mathrm{ref}}}\left( s \right) \right\} \right] 
\\
\,\,\,\,          &=V^{\pi _{\mathrm{ref}}}\left( s \right) 
\\
\,\,\,\,          &=V^{\pi _{\mathrm{ref}}}_\beta\left( s \right) 
\end{align*}
which means $V_{\beta}^{\pi _{\mathrm{ref}}}$ is the fixed point of the Bellman optimality operator thus $\pi_{\text{ref}}$ is the optimal policy by Theorem \ref{theorem:optimal policy}

\end{proof}

\end{document}